\documentclass[conference]{IEEEtran}
\usepackage{times}

\usepackage[numbers]{natbib}
\usepackage{multicol}

\usepackage{multirow}

\usepackage{mathtools}
\usepackage{amsthm}
\usepackage{graphicx}
\usepackage{amsmath}
\usepackage{amssymb}
\usepackage{booktabs}
\usepackage{array}
\usepackage{wrapfig}
\usepackage{multirow}

\usepackage[pagebackref=false,breaklinks=true,colorlinks,bookmarks=false,urlcolor=blue, citecolor=blue]{hyperref}

\DeclareMathAlphabet{\pazocal}{OMS}{zplm}{m}{n}

\DeclareMathAlphabet\mathbfcal{OMS}{cmsy}{b}{n}

\begin{document}

\title{Learning 3D Semantics from Pose-Noisy 2D Images with Hierarchical Full Attention Network}


\author{\authorblockN{Yuhang He}
\authorblockA{Department of Computer Science\\
University of Oxford\\
Oxford, UK\\
Email: yuhang.he@cs.ox.ac.uk}
\and
\authorblockN{Lin Chen}
\authorblockA{Institute of Photogrammetry and \\
GeoInformation, Leibniz University Hannover\\
Hannover, Germany\\
Email:  chen@ipi.uni-hannover.de}
\and
\authorblockN{Junkun Xie and Long Chen}
\authorblockA{School of Computer Science and Engineering\\
Sun Yat-Sen University, China.\\
xiejk3@mail2.sysu.edu.cn\\
long.chen@ia.ac.cn}}


%

\maketitle

\begin{abstract}
We propose a novel framework to learn 3D point cloud semantics from 2D multi-view image observations containing pose error. On the one hand, directly learning from the massive, unstructured and unordered 3D point cloud is computationally and algorithmically more difficult than learning from compactly-organized and context-rich 2D RGB images. On the other hand, both LiDAR point cloud and RGB images are captured in standard automated-driving datasets. This motivates us to conduct a ``task transfer'' paradigm so that 3D semantic segmentation benefits from aggregating 2D semantic cues, albeit pose noises are contained in 2D image observations. Among all difficulties, pose noise and erroneous prediction from 2D semantic segmentation approaches are the main challenges for the ``task transfer''. To alleviate the influence of those factor, we perceive each 3D point using multi-view images and for each single image a patch observation is associated. Moreover, the semantic labels of a block of neighboring 3D points are predicted simultaneously, enabling us to exploit the point structure prior to further improve the performance. A hierarchical full attention network~(HiFANet) is designed to sequentially aggregates patch, bag-of-frames and inter-point semantic cues, with hierarchical attention mechanism tailored for different level of semantic cues. Also, each preceding attention block largely reduces the feature size before feeding to the next attention block, making our framework slim. Experiment results on Semantic-KITTI show that the proposed framework outperforms existing 3D point cloud based methods significantly, it requires much less training data and exhibits tolerance to pose noise. The code is available at~\url{https://github.com/yuhanghe01/HiFANet}.
\end{abstract}

\IEEEpeerreviewmaketitle

\section{Introduction}
Directly learning from 3D point cloud is difficult. Challenges derive from four main aspects: First, 3D point cloud is massive and a typical Velodyne HDL-64E scan leads to millions of points. Processing such large data is prohibitively expensive for many algorithms and computation sources. Second, 3D point cloud is unstructured and unordered as well. It record neither the physical 3D world texture nor object topology information, which have often been used as important priors by image based environment perception methods~\cite{SSD,faster_rcnn,resnet}. Third, data imbalance issue. Due to the 3D physical world layout that particular categories conquer most of the space, captured 3D point cloud is often dominated by classes such as road, building and sidewalk. Other categories\,(\textit{i.e.} traffic sign, poles, pedestrian) with minor point cloud presence but vital importance for self-driving driving scenario understanding and high-quality map construction are often overwhelmed by dominating classes. Lastly, capturing 3D point cloud is a dynamic process, resulting in inconsistent and nonuniform data sampling. Distant objects are much more sparsely sampled than close objects.

The aforementioned difficulties largely restricted 3D point cloud segmentation progress. 3D point cloud processing with deep neural network~\cite{pointnet,pointnet_pp,RandLA-Net,bo_net,SPG} has thus emerged much later than counterpart task in 2D images~\cite{cityscapes,kitti_dataset,faster_rcnn,mscoco,imagenet_cvpr09,resnet}. Meanwhile, most self-driving data collection platforms collect 3D point cloud and RGB images simultaneously, with the LiDAR scanner and camera being pre-calibrated and synchronized to perceive the scene. Therefore, we are naturally motivated to transfer 3D point cloud segmentation to its 2D image based counterpart\,(we call ``task transfer'') so that the segmentation of point cloud can largely benefit from various matured 2D image semantic segmentation networks. Specifically, we exploit features arising from 2D image semantic segmentation result to predict 3D point cloud semantics.

The feasibility of such ``task transfer'' basically lies in the fact that, given the LiDAR-Camera pose, we can project a 3D point to the 2D image plane to get its 2D pixel correspondence. However, such seemly-fascinating ``task transfer'' comes with a price: In real-scenario, LiDAR-Camera pose is often noisy so accurate 3D-2D correspondences are non-guaranteed. In addition, view-angle change easily results in distorted image observation. Moreover, 2D semantic segmentation method may also give erroneous predictions. 

To tackle the aforementioned challenges, we first propose to perceive each 3D point from multi-view images so that bag-of-frame observations for each single 3D point are obtained. Multi-view image observation reduces the impact of the unfavoured view-angle as it introduces extra semantic cues. Moreover, instead of looking into single-pixel of an image, we focus on a small patch-area around the pixel. The patch observation strategy mitigates 3D-2D correspondence error led by pose noise and further enables neural network to learn pose noise tolerant representation in a data-driven way. Moreover, we process a local group of spatially or temporally close 3D points at the same time, so that we can exploit 3D points structure prior\,(\textit{i.e.} two points' spatial location). Actually, the local 3D point group and the corresponding 2D observation can be treated as seq2seq learning problem~\cite{seq2seq_nips14}, where one sequence is 2D image and the other is 3D point cloud. To accommodate these different data representation properties, we propose a hierarchical fully attention network~(HiFANet) to sequentially and hierarchically aggregate the patch observation, bag-of-frame observation and inter-point structural prior to infer the 3D semantics. Such hierarchical attention blocks design enables the neural network to learn to efficiently aggregate semantics at different levels. Moreover, the preceding attention block naturally reduces the feature representation size before feeding it to the next attention block, so the whole framework is slim by design.


In sum, our contribution is three fold: first, we propose to transfer 3D point semantic segmentation problem to its counterpart in 2D images. Second, to counteract the pose noise impact, we propose to associate each single 3D point with multi-view patch observation so that the neural network can learn to tolerate pose inaccuracy. Third, we formulate it as a seq2seq problem so that we can best exploit the structural prior arising from both 3D point cloud and 2D images to improve the performance. 

\section{Related Work}
\label{sec:relatedwork}

3D semantic segmentation can be divided into three main categories: point-based, voxel-based and 2D projection based methods~\cite{he2021deep,xie2020linking}.

Point based methods compute the features from points and can be categorized into three sub-classes~\cite{he2021deep}: Multi Layer Perceptron (MLP), point convolution and graph convolution based methods.  MLP based method apply MLP directly on points to learn features, such as PointNet ~\cite{pointnet}, HRNN~\cite{ye20183d}, PointNet++~\cite{pointnet_pp}, PointWeb~\cite{zhao2019pointweb}. In comparison, point covolution based methods apply convolution on individual point. Representative works in this group are PointwiseCNN~\cite{hua2018pointwise}, PCNN~\cite{wang2018deep}, PointConv~\cite{wu2019pointconv}, RandLA-Net~\cite{hu2020randla} and PolarNet~\cite{zhang2020polarnet}. In the third class, the points are connected with graph structure, graph convolution is further applied to capture more meaningful local information. Example works include DeepGCNs~\cite{li2019deepgcns}, AGCN~\cite{xie2020point}, HDGCN~\cite{liang2019hierarchical} and 3DContextNet~\cite{zeng20183dcontextnet}. 

In voxel based methods, voxels divide 3D space into volumetric grids, which are used as input for 3D convolutional neural networks. The voxel used is either uniform~\cite{huang2016point,dai2018scancomplete,meng2019vv} or non-uniform~\cite{riegler2017octnet,graham20183d}. Methods in this group are restricted by the fact that the computation burden fast grows with the scale of scene. Consequently, the usage of those methods in large scale becomes impractical.

In projection based methods, point cloud is projected into synthetic but multi-view image planes and then 2D CNNs are used by each view, finally semantic results from mutliple views are aggregated~\cite{lawin2017deep,guerry2017snapnet,transform3D2D,sparse_upsampling},  However, this idea is restricted by misinterpretation stem from sparse sampling of 3D points. Our work shares the similar idea to convert 3D point cloud to 2D plane, but we exploit 2D RGB images to assist 3D semantic segmentation and we rely on 2D semantic segmentation to predict 3D semantics.

In 2D semantic segmentation, FCN~\cite{long2015fully} is one of the first works using deep neural network for semantic segmentation by replacing the fully connected layer with fully convolution layers. The following works, e.g., SegNet~\cite{badrinarayanan2017segnet} and~\cite{noh2015learning}, use more sophisticated way to encode the input image and decode the latent representation so that images are better segmented. Obtaining features at multiple scale is manipulated either at convolution kernel level or through pyramid structure. The former leads to the method of using dilated convolution and representative works are DeepLabV2~\cite{chen2017deeplab} and~DeepLabV3\cite{chen2018encoder}. The latter is implemented in PSPN~\cite{zhao2017pyramid} and~\cite{ghiasi2016laplacian}. Also, attention mechanisms are used to weight features softy for semantic segmentation task in~\cite{chen2016attention}. In this paper, we make use of the network proposed in~\cite{semantic_video_cvpr19} as our base feature extractor, since it uses synthetic predicting to scale up training data and the trained label is also robust, benefiting from the usage of the boundary relaxation strategy proposed in that paper.

This paper utilizes features from multi-view patches sampled from camera images, which are not accurately aligned with 3D point cloud, to benefit the semantic segmentation of 3D point cloud. In this context, the central issue is how to aggregate multi view image features in a sophisticated way so that 3D points can be better separated in the feature space spanned by those aggregated features.   


\section{Problem Formulation}
We have a sequence of $N$ 3D point cloud frames $\mathbf{P} = \{P_1, P_2, \cdots, P_N\}$, and framewise associated 3D point semantic label $\mathbf{C}$ and RGB image $\mathbf{I}$. Such data is collected by platform where LiDAR scanner and camera are carefully synchronized and pre-calibrated with noisy pose information $P_{o}=[R|t]$\,(rotation matrix $R$ and translation $t$). Moreover, the relative pose between any two neighboring point cloud frames can be obtained via IMU system. With the noisy pose, we can theoretically project any 3D point to any image plane. Off-the-shelf image semantic segmentation method~\cite{semantic_video_cvpr19} is adopted to get semantic result $\mathbf{S}$ for each image, each pixel of which consists of categorical semantic label and semantic-aware representation $r$. Our goal is to train a model $\mathbfcal{F}$ parameterized by $\theta$ to predict point cloud semantics from images $\mathbf{C} = \mathbfcal{F}(\mathbf{I},\mathbf{S}|P_{o}, \theta)$.



\section{Hierarchical Full Attention Network}

\begin{figure*}[t]
    \centering
    \includegraphics[width=0.95\linewidth]{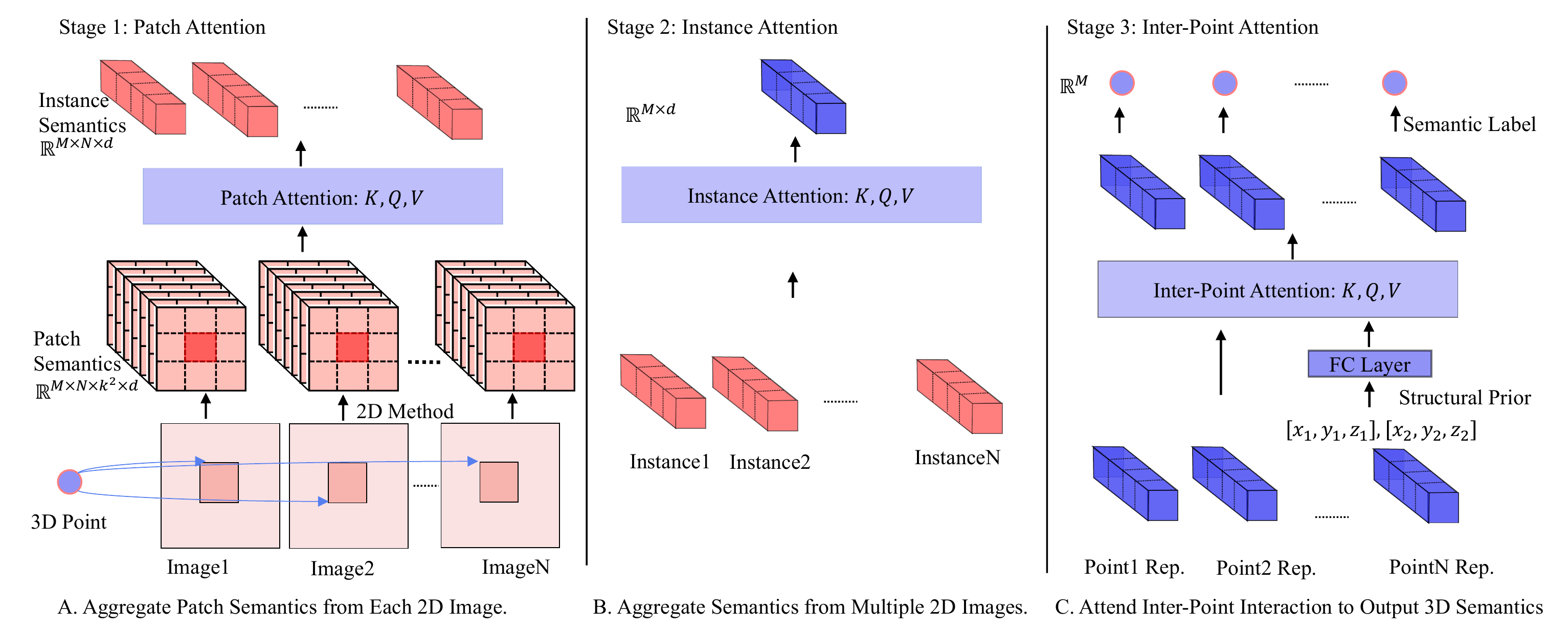}
    \caption{HiFANet pipeline: Given the pose, we project $M$ 3D points to their nearest top-N RGB images to get $k\times k$ patch observations. Off-the-shelf 2D image segmentation model is trained to get each patch's semantic feature representation as well as categorical semantic labels. HiFANet is a three-stage hierarchical fully attentive network. It first learns to aggregate patch representation into an instance representation\,(left image), then aggregates multiple image instances into one point-wise representation\,(middle image), and finally an inter-point attention module to attend structural and feature interaction among 3D points to output per-point elegant semantic feature representation, which is then used to predict the ultimate semantic label.}
    \label{fig:pipeline}
\end{figure*}

The fundamental idea of designing our framework is two-fold: ``task transfer'' which learns 3D point cloud semantics from 2D images; further address accompanying challenges brought by the ``task transfer'' through a ``learning'' perspective by fully exploiting the potential of deep neural networks in a hierarchical way. Specifically, given the pose information between any 3D point cloud frame and any 2D image, we can obtain $N$ patch observations $\{\mathcal{P}_1,\cdots, \mathcal{P}_N\}$ for each 3D point by projecting it to its neighboring image frames\,(we call bag-of-frames), where a patch observation $\mathcal{P}_i$ indicates a $k\times k$ squared patch centered at the pixel $[u_x, u_y]$ of the 3D point's $i$-th observation image frame. $[u_x, u_y]$ corresponds to the 3D point projection location with noisy pose information. In the meantime, a pre-trained 2D image semantic segmentation model is available, so we can get both the categorical semantic label $s_j$ and the semantic-guided feature representation $r_j$ for the $j$-th pixel in the patch. The feature representation $r_j$ can be easily obtained by taking the penultimate layer activation of the model trained on 2D images. So the patch observation can be expressed as,
\begin{equation}
    \mathcal{P} = \{(s_1, r_1)_i, \cdots, (s_{k^2}, r_{k^2})_i\}_{i=1}^{N}
\end{equation}

Introducing patch observation instead of single-pixel observation in 2D image is to address pose noise challenge, which we will give detailed discussion in next section. Instead of learning 3D semantic for each 3D point separately, we model $M$ neighboring 3D points simultaneously, which benefits us to use 3D points structure prior to escalate the performance. For example, an intuitive spatial prior is that two spatially-close 3D points are much more likely to share the same semantic label than those lie far apart. In sum, our model takes $M$ 3D points' Cartesian coordinates as well as each 3D point's $N$ patch observations as input and outputs each 3D point's semantic label. It is worth noting that $M$ 3D points forms a point sequence and $M\times N$ image observation forms another image sequence, the whole framework can be treated as a seq2seq task, either spatially or temporally. The framework input simply consists of image-learned semantic information\,(categorical label or feature presentation), no extra constraint is involved and we do not directly process 3D point cloud.


With ``task transfer'', the main task of our framework is to efficiently aggregate semantic clues arising from bag of 2D image frames. To this end, we propose a hierarchical full attention three-stage aggregation mechanism, in which we first learn to aggregate patch observation into an instance observation\,(i.e., single pixel observation in an image), and then learn to aggregate multiple instances in the bag-of-frames for each 3D point into 3D point wise observation, and finally attend all the structure prior and interaction between 3D points to output the target semantic label for each single 3D point. Our framework is fully attentive and invariant to images observation order permutation. The hierarchical attention mechanism design has two advantages: it first enables the neural network to fully learn specified attention tailed for different semantic representation, second it aggressively reduces the feature size so that we keep the whole framework slim.




\subsection{Patch Attention for Patch Aggregation} 
Patch attention tends to aggregate the patch observation into a single-pixel observation. Within each $k\times k$ patch, we call the centered point the principle point and the remaining points are neighboring points. The basic idea behind the patch attention is to attend all points in the patch with a trainable weight before weighted-adding them together to generate one feature. Since the principle point records the most-confident 3D point semantic related feature representation, we add a short-cut connection between the principle point feature and the attended to feature representation. Specifically, given the feature representation $r_i \in \mathbb{R}^{k\times k \times d}$, the principle points lies in $[\frac{k}{2}, \frac{k}{2}]$ and has feature representation $f_p$ of length $d$, the output feature $f_{pa}$ after patch attention can be expressed as,

\begin{equation}
    f_{pa} = \sum_{j=1}^{k\times k} w_j\cdot V_j + f_p
\end{equation}

where $w_j$ is the learned weight for the $j$-th point in the patch. To learn the attention weight $w$, we draw inspiration from self-attention module~\cite{att_all_need} to learn a patch Key $K = \mathbb{R}^{k\times k \times d_1}$ and patch Query $Q = \mathbb{R}^{k\times k \times d_1}$ and a patch Value $V = \mathbb{R}^{k\times k \times d}$. The three parts can be efficiently learned via $1\times 1$ 2D convolution on the patch observation. To reduce the computation cost (usually $d_1\ll d$), we set $d_1 = 64$ and $d = 256$. With $K$ and $Q$ we can further compute the scaled dot-product attention where the attended weight $w$ can be obtained by,

\begin{equation}
    w = softmax(\frac{Q_p K}{\sqrt{d_1}})
\label{eqn:attention}
\end{equation}

$Q_p$ is the principle point query. With Eqn.~(\ref{eqn:attention}), we can get the weight of each point to the principle point. The patch attention is a self-attention module, it requires no extra supervision and can efficiently attend the final single-pixel observation in with paralleling computation.

\subsection{Instance Attention for Image Aggregation}
Instance attention module takes $\mathbb{R}^{M\times N \times d}$ semantic feature as input, and aims to aggregate bag-of-frames features to get 3D point wise feature. We call the aforementioned patch-attention aggregated pixel-wise semantic representation in each image frame as an instance, because it represents an independent observation towards a 3D point. The multiple instances arising from bag-of-frames form an \textit{Instance Set}~\cite{five_instance,lee2019set}, which means these instances are orderless, the final accurate semantic label may derive from an individual instance or multiple instances combination. To satisfy the instance set property, the instance attention module has to be order-permutation invariant. Commonly seen set-operators include max-pooling and average-pooling. In HiFANet, we first apply a self-attention layer like the patch attention block does to attend each instance by all the remaining instances. Finally, we apply average pooling to merge multiple instances into one instance representation.


\subsection{Inter-point Attention for 3D Points Aggregation}
Inter-point attention take $\mathbb{R}^{M\times d}$ semantic feature learned by instance attention module as input. Unlike the previous two attention modules that just focus on per-point semantic feature learning, inter-point attention module fully considers the interaction between 3D points, including the spatial structure interaction and semantic feature interaction. We adopt a Transformer~\cite{att_all_need} multi-head self-attention like network to construct the inter-point attention module. Specifically, the input feature is fed to learn per-point Key $K=\mathbb{R}^{M\times d_2}$ and per-point Query $Q=\mathbb{R}^{M\times d_2}$ as well as per-point Value $V = \mathbb{R}^{M\times d}$. To involve structural prior, we encode the relative Cartesian position difference between any two 3D points $p_i - p_j$. The Cartesian position difference is further fed to two consecutive fully connection layers to get the structural prior encoding $K_{pe}$, which is the same size of $K$. The original Key $K$ is then updated by adding $K_{pe}$,

\begin{equation}
    K = K + K_{pe}
\label{update_K}
\end{equation}

The updated $K$ in Eqn.(\ref{update_K}) naturally contains the structural prior. With the $Q$ and updated $K$, we can compute the attention weight for each single 3D point w.r.t the remaining 3D points, as is shown in Eqn.\,(\ref{eqn:attention}). The attention weight is further applied to combine value $V$ to get the final per-point semantic representation, which is further concatenated with a classification layer for semantic classification.

In sum, HiFANet sequentially and hierarchically aggregates patch semantics, instance semantics and inter-point semantics to learn semantic representation for each 3D point. It is fully attentive and learns compartmentalized and certain attention blocks w.r.t. different aggregation granularity separately. The preceding attention layer largely reduce the feature size before feeding it to the next layer, so the whole neural network is slim. Detailed HiFANet pipeline is shown in Fig.\,\ref{fig:pipeline}.


\section{Discussion on HiFANet Design Motivation}
The feasibility of such ``task transfer'' lies in the availability of the pose information between LiDAR scanner and the camera, which enables us to project 3D point cloud onto the image plane to get each 3D point's correspondence in the image plane. We hereafter call such correspondence as a 2D observation. The ``task transfer'' poses three main challenges that may jeopardize the performance.

\begin{enumerate}
    \item Pose noise. Sensor calibration often suffers from internal and external noise. Noisy pose information leads to inaccurate 2D observations. This stays as the most prominent challenge.
    \item View-angle. Projecting a cluster of point cloud belonging to a specific category\,(\textit{i.e.} car) to an image plane often leads to distorted 2D observation. In severe cases, it leads to wrong observation due to the occlusion caused by view-angle difference.
    \item Void projection. While LiDAR scanner scans in $360^\circ$, pinhole camera simply captures the forward-facing view. This mismatch of perception field inevitably leads to void projection in which point cloud cannot find observation in one image.
\end{enumerate}

Addressing the above three challenges leads to our proposed framework. To mitigate the pose noise impact, we propose to use patch observation to replace pixel observation. Pixel-wise observation is fragile and sensitive to pose noise, a small change leads to totally different observation. Patch-wise, on the contrary, becomes much more resilient to pose noise because it covers possible observations potentially led by noisy pose. Moreover, introducing patch-wise observation avoids us directly optimizing $[R|t]$ in an iterative way. To address the view-angle and void projection issue, we propose to involve multiple observation arising from different view-angles. With the multi-view observations, we naturally obtain multiple clues for each 3D point.

\subsection{Pose Noise and Patch Observation}
\label{sec:pose_discuss}
The pose between LiDAR scanner coordinate system and camera coordinate system can be formulated as a rotation matrix $R$ and translation $T$. A 3D point $[x,y,z]$ projects onto an image plane, the corresponding observation location $[u, v]$ in the 2D image plane is computed by,

\begin{equation}
    [u, v, 1] = K[R|T] \cdot [x, y, z, 1]^T  
    \label{eqn:projection}
\end{equation}

Please note that the projected pixel location is normalized by its 3rd dimension. The pose noise of sensor calibration (between laser scanner and camera) renders the location of true projected point uncertain. However, in our approach, a patch is extracted and then the attention is learned to focus on the pixel closest to the true projected points. 

\begin{figure}
    \centering
    \includegraphics[scale=0.27]{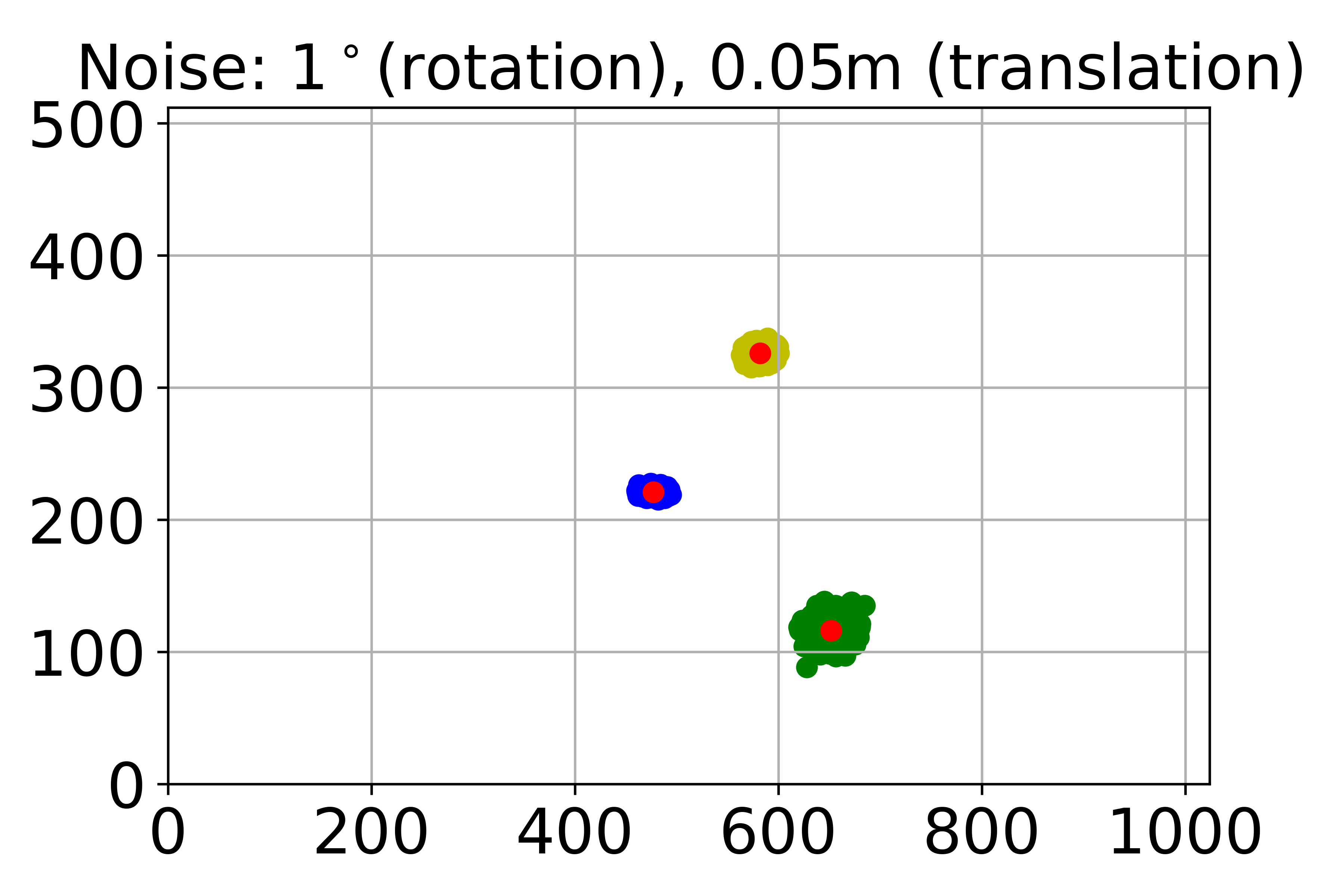}
    \includegraphics[scale=0.27]{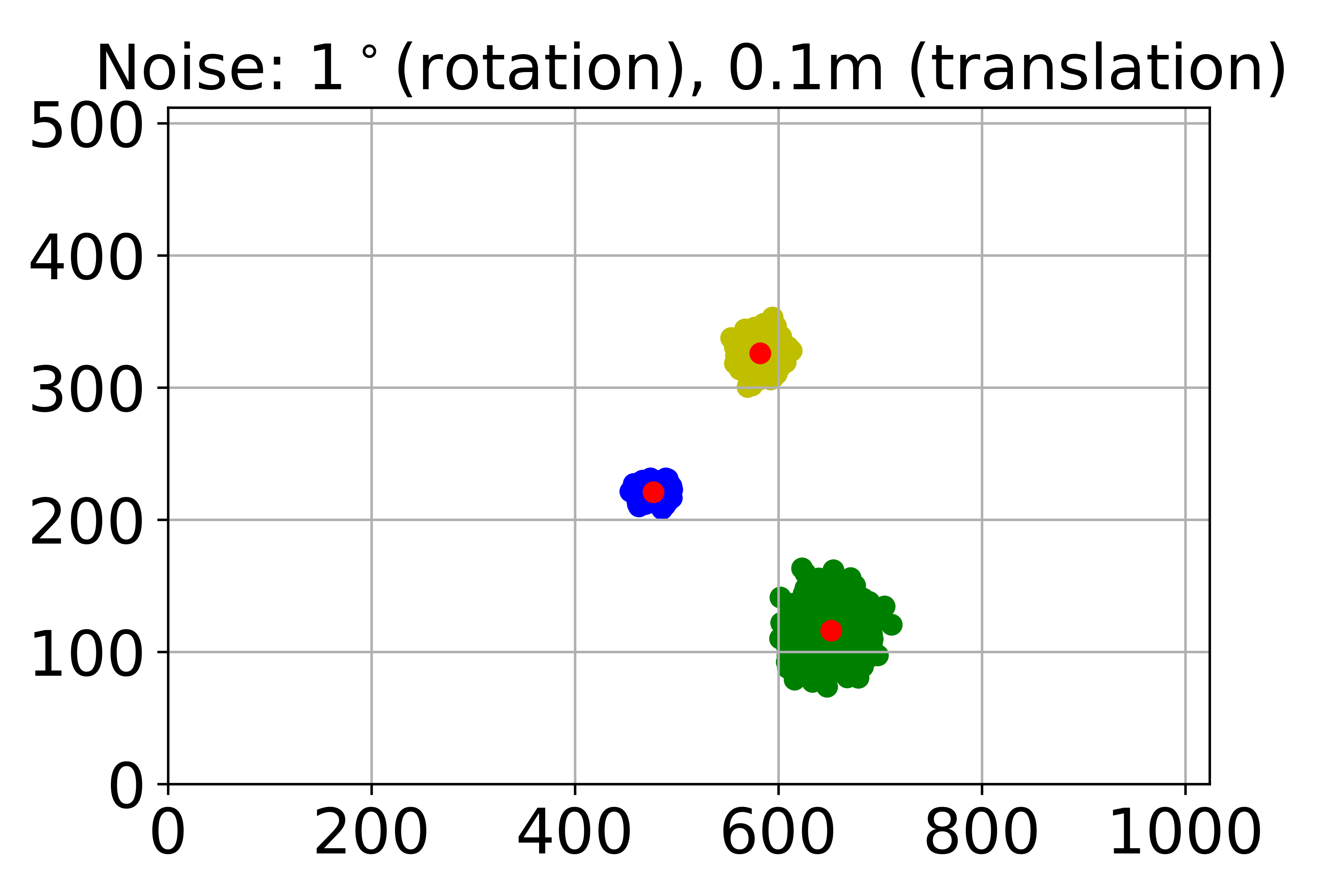}
    \caption{The influences of pose noise to the projected point coordinates on image plane (in pixel) for 3D world points that are various distant from the camera plane (green: 5m; yellow: 10m; blue:20m). The simulated noises of rotation angles are $1^\circ$ (for each rotation angle) for both cases and the translation noise are 0.05m (left), 0.1m (right) for each of the three axes in world coordinate system.}
    \label{fig:my_label}
\end{figure}

In order to investigate the influence of the pose noise on the location of projected points, a toy simulation experiment is provided and illustrated in Fig.~\ref{fig:my_label}. As can be observed in Fig.~\ref{fig:my_label}, given the translation noise for the calibration between the LiDAR scanner and camera as 10cm and the rotation angle noise as 1$^\circ$, the projection error on the image plane (1024 $\times$ 512 pixels) is around 40 pixels for near camera object points. Since the patch extracted on each camera view is within a $k\times k$ patch in the downsized feature maps (normally at $1/16$ or $1/32$ resolution), the information encoded in the image is then well preserved for the attention module to discover, although the pose noise exist. 

\subsection{View-angle and Void Projection}
View-angle easily leads to titled, occluded and even erroneous observation. A 3D point that is observed in one viewpoint~(an RGB image) can be obstructed in another neighboring viewpoint. Traditional 3D reconstruction framework like structure-from-motion\,(SfM~\cite{pmvs}) suffer from the same dilemma. The void projection jeopardizes the ``task transfer'' proposal because it causes large number of 3D points being 2D image unobserved. 


To mitigate the two challenges, we propose to observe a single 3D point from multiple view-angles. On the one hand, it reduces the risk of one 3D point being observed at an unfavored view angle. On the other hand, it maximally ensures each 3D point cloud to be observed by at least one 2D image. Moreover, this strategy brings us the advantage of aggregating semantic clues arising from multiple images to better estimate semantics. Multiple view-angles observation can be efficiently aggregated in parallel in HiFANet.

\section{Experiments and Results}

\begin{table*}[htbp]
\small
\caption{Quantitative Result on Semantic-KITTI\cite{semantic_kitti} Dataset. B, K and M mean billion, thousand and million, respectively.}
\begin{center}
    \begin{tabular}{ |c|c|c|c|c|c| } 
 \hline
 Method Category & Method & Train Dataset & Param Num & mIoU ($\uparrow$) & Average Accuracy ($\uparrow$) \\ 
 \hline
 \multirow{7}{*}{Point Based Methods} & PointNet~\cite{pointnet} & 2.8 B & 3.53 M & 0.036 & 0.105\\ 
 &PointNet++~\cite{pointnet_pp} & 2.8 B &  0.97 M & 0.055 & 0.156 \\ 
 &RangeNet++(CRF)~\cite{milioto2019iros} & 2.8 B & 50.38 M & 0.500 & 0.878 \\ 
 &RangeNet++(KNN)~\cite{milioto2019iros} & 2.8 B & 50.38 M & 0.512 & 0.899 \\ 
 &KPConv\cite{KPConv} & 2.8 B & 18.34 M & 0.466 & 0.868 \\ 
 &RandLANet~\cite{RandLA-Net} & 2.8 B & 1.24 M & 0.578 & 0.913 \\
 \hline
 \multirow{6}{*}{Image Aggregation Methods} & BoF Num = 1 & 23 K & 137 M & 0.422 & 0.845 \\ 
 &BoF Num = 3 & 23 K & 137 M &0.437 & 0.852 \\ 
 &BoF Num = 5 & 23 K & 137 M &0.436 & 0.852 \\ 
 &Patch Size = 1 & 23 K & 137 M &0.436 & 0.850 \\ 
 &Patch Size = 3 & 23 K & 137 M &0.436 & 0.851 \\ 
 &Patch Size = 5 & 23 K & 137 M &0.436 & 0.852 \\ 
 \hline
\multirow{2}{*}{Multi-View Learning Methods} & AvgPool\_FC & 0.5 M & 0.04 M & 0.451 & 0.872 \\ 
&HiFANet\_noPA & 0.5 M & 2.5 M &0.537 & 0.891 \\
&HiFANet\_noSP & 0.5 M & 2.7 M &0.561 & 0.920 \\
 &HiFANet & 0.5 M & 2.7 M &\textbf{0.620} & \textbf{0.933} \\ 
 \hline
\end{tabular}
\label{fig:quant-rst}
\end{center}
\end{table*}

\label{sec:experiment}
We conduct experiment on the Semantic-KITTI dataset~\cite{semantic_kitti}. Since we need the inter-frame odometry information to project each 3D point to multiple RGB frames but the official provided test dataset (sequence 11-20) does not provide such information, so we do not follow the official split but instead create the train/test/val split by ourselves and further train the comparing methods with the split dataset from scratch. The same problem applies to other relevant datasets such as Waymo and CityScapes~\cite{cityscapes}, so we just run experiment on Semantic-KITTI dataset in this paper. 


\begin{figure*}[t]
    \centering
    \includegraphics[width=0.98\linewidth]{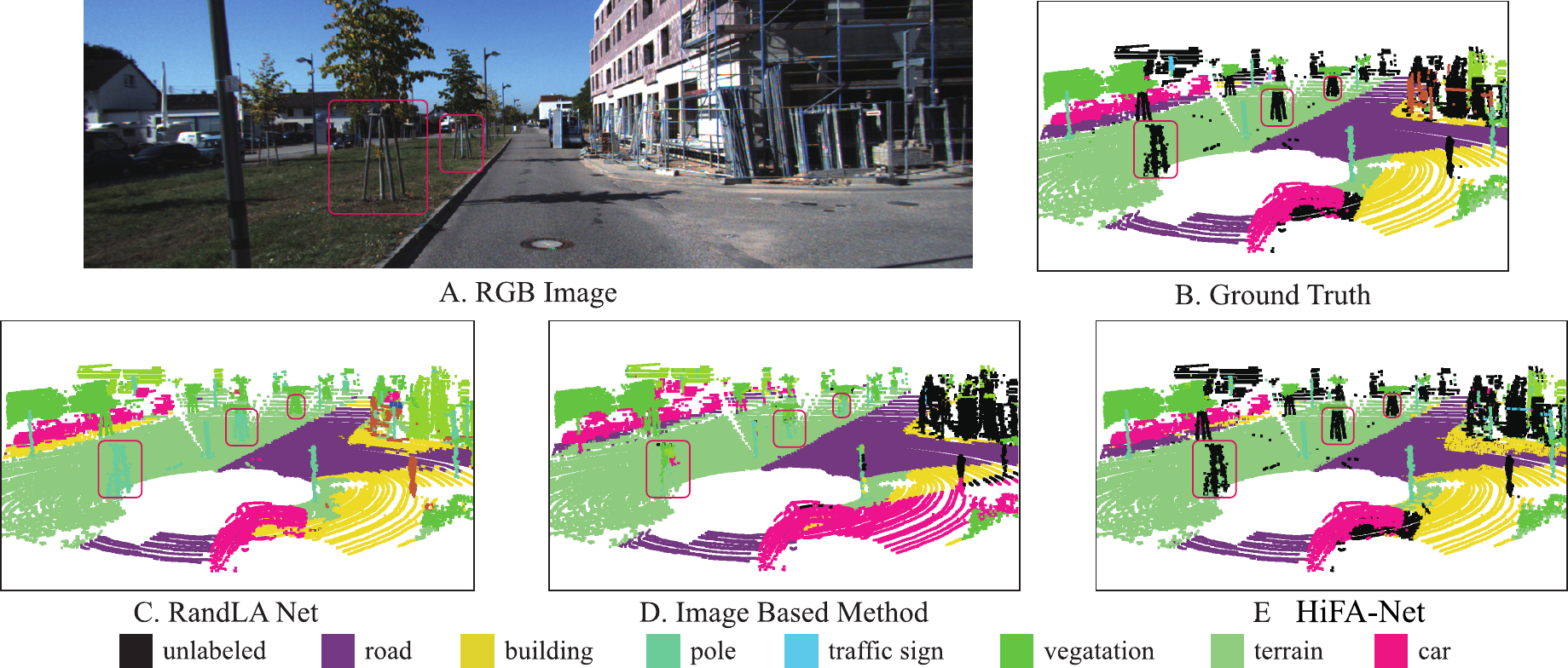}
    \caption{Close-up visualization of various methods on unlabelled tree stake. While point based method erroneously classifies them as pole and image based method as terrain, HiFANet accurately recognizes it by fully combing 2D image based semantics and 3D structural priors.}
    \label{fig:quantiative_tree_pole}
\end{figure*}

\begin{figure*}[t]
    \centering
    \includegraphics[width=0.98\linewidth]{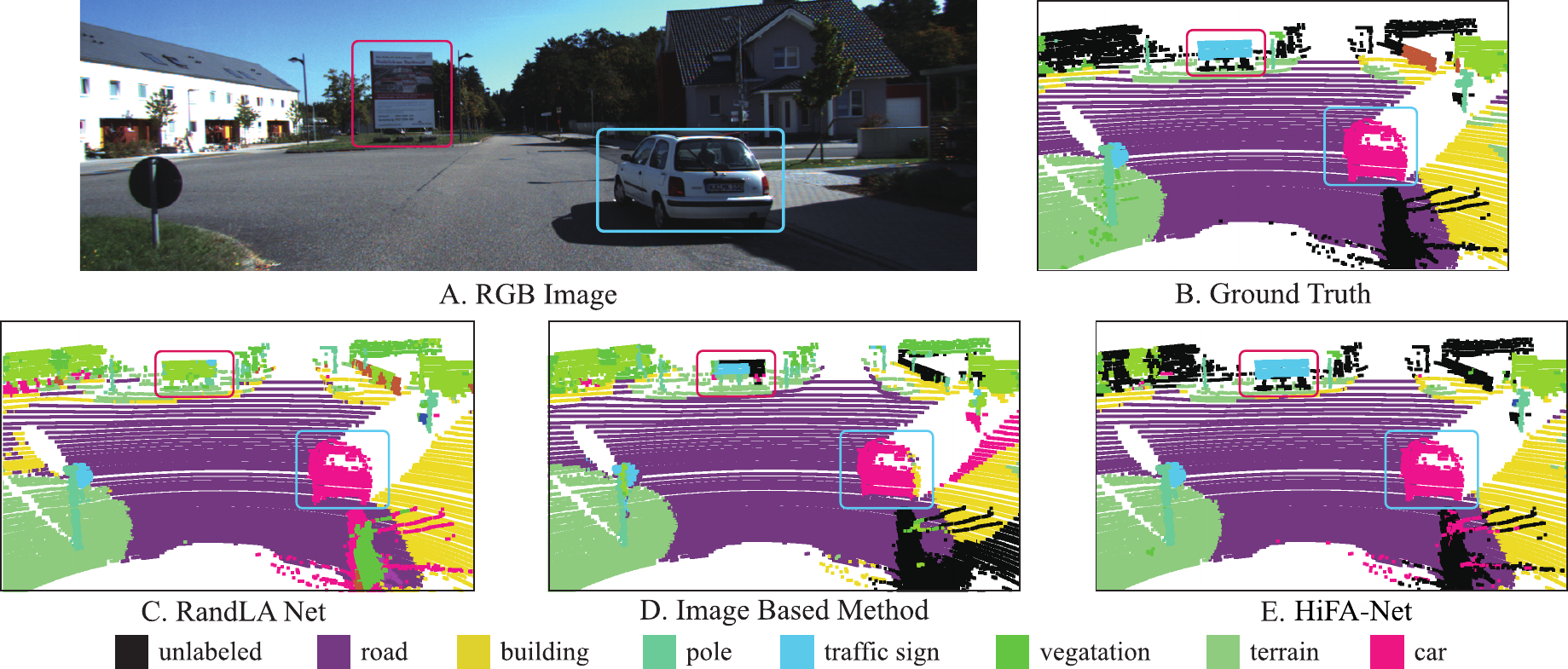}
    \caption{Global visualization of various methods comparison. While point based method fails to classify traffic sign and image based method generates spatially distributed prediction, HiFANet successfully avoids these dilemmas and gives the right semantics.}
    \label{fig:quantiative_global}
\end{figure*}

\textbf{Data Preparation} We run experiment on sequence 00-10 because the inter-frame odometry information is available for the 11 sequences, with which we can register all point cloud frames from a sequence to a uniform 3D coordinate so that each 3D point can be freely projected to any image plane. There are 13 semantic categories in total: \textit{road}, \textit{side-walk}, \textit{building}, \textit{fence}, \textit{pole}, \textit{traffic sign}, \textit{vegetation}, \textit{terrain}, \textit{person}, \textit{bicyclist}, \textit{car}, \textit{motorcycle} and \textit{bicycle}. Some categories like \textit{road}, \textit{building}, \textit{vegetation} and \textit{terrain} dominate most of the points, whereas the others' portion is very small. An extra \textit{unlabelled} background category is added. Sequence 06 is selected as test set as it contains all semantic categories and account for $20\%$ data of the whole dataset. Sequence 08 is selected for validation and the remaining 9 sequences serve as training set. To get each 3D point's $N$ neighboring image observations, we project it to its closest N image planes. $N$ is set as 5 because it then covers $64\%$ of the whole point cloud dataset with patch size $k = 5$ and 3D points number size $M=10$. Those 3D points that fail to find $N$ image observations are discarded during test but left for training point cloud based models. The image based semantic representation and semantic label are obtained from VideoProp~\cite{semantic_video_cvpr19} model pre-trained on KITTI dataset~\cite{kitti_dataset}. The semantic representation is a 256-d feature. Therefore, the size of patch semantics representation feeds to HiFANet is $5\times 5 \times 256$. For the evaluation metric, we adopt the standard mIoU and average accuracy~\cite{semantic_kitti}.

\textbf{Methods to Compare} The first method category we tend to compare is pure 3D point cloud based semantic segmentation method. It helps us to gain an understanding of how far our proposed ``task transfer'' strategy goes, comparing with directly learning from 3D points. The second method category we compare with is the semantic result giving by deterministically aggregating the category semantic labels predicted by 2D image aggregation method, it gives us an understanding of how good image based semantic prediction methods can perform, by varying the observation number like image number and patch size. The third category is multi-view learning method which means designing neural network to learn from image semantic representations, as our proposed HiFANet does.

\textbf{Ablation Study} we want to figure out the impact of the involvement of patch feature representation, structural prior on the performance. We thus test two HiFANet variants: reduce the patch size to 1 so no patch attention module is applied\,(\textbf{HiFANet\_noPA}), no structural prior involvement in inter-point attention module\,(\textbf{HiFANet\_noSP}). Moreover, to test the effectiveness of our proposed full attention network, we train another simple semantic aggregation network, in which we simply average-pool all the input feature\,(patch and instance feature) to get per-point feature, and further concatenate two full connection layer\,(of size 256, 128) to directly predict the semantic label\,(\textbf{AvgPool\_FC}). Please note that AvgPool\_FC is a simple neural network and it is order-permutation invariant.

Five most recent 3D point cloud based methods: PointNet~\cite{pointnet}, PointNet++~\cite{pointnet_pp}, RangeNet~\cite{milioto2019iros}\,(two variants, with KNN and CRF), KPConv~\cite{KPConv} and RandLANet~\cite{RandLA-Net} are selected for comparison study. For image aggregation methods, we simply deterministically choose the semantic label with maximum occurrence times. Within multi-view learning methods, all HiFANet variants are trained with the same hyper-parameter setting as HiFANet.

\textbf{Quantitative Result} is shown in Table\,\ref{fig:quant-rst}. We can observe that point cloud based methods training requires much larger number of training dataset than both image aggregation methods and our proposed multi-view learning methods. This shows the advantage of learning semantics from 2D images. The compactly-organized and topology-preserving RGB images enables neural network to learn meaningful semantic representations with much fewer training samples. Within image aggregation methods, involving extra bag-of-frame observations increases the performance, but the performance gain is not prominent due to the view-angle and occlusion challenges. Moreover, expanding the patch size also improves the performance, which shows capability of introducing patch-wise observation in mitigating the dilemma caused by observation uncertainty. In sum,  aggregating image-predicted semantics can achieve comparable performance than point based methods. It further shows the potential of designing neural network to learn from image learned semantic representations, instead of simply voting them.

Within multi-view learning methods, we can observe that all methods outperform image aggregation methods, showing the advantage of neural network learning over deterministic semantic aggregation. Simply adding several fully connection layers\,(AvgPool\_FC) generates inferior performance than the other three HiFANet variants. This result shows that more advanced semantic aggregation strategy is needed to better aggregate semantic cues arising from multiple image observations. At the same time, either removing the patch attention module or the structural prior module inevitably reduces the performance. Patch observation introduces extra semantic cues in a pose noise sensitive way and structural prior regularizes the whole network training. Finally, HiFANet generates the best performance over all methods, far outweighing other methods by a large margin. 

\textbf{Qualitative Result} is shown in Fig.\,\ref{fig:quantiative_tree_pole} and Fig.\,\ref{fig:quantiative_global}. In the close-up comparison of tree stakes in Fig.\,\ref{fig:quantiative_tree_pole}, as it is a category falls out of our consideration, it should be regarded as \textit{unlabelled} category. However, 3D point based method RandLANet~\cite{RandLA-Net}\,(sub-figure B.) mixes it with \textit{pole} due to their point cloud representation similarity. Image aggregation method\,(sub-figure D.) directly predicts it as terrain because of its color similarity and connection with the tree leaves. HiFANet\,(sub-figure E.), however, fully exploits 3D point structural prior information to predict the correct semantics. For example, the tilted angle of tree stakes over the ground makes it unlikely to be a pole\,(which is usually vertical to ground), nor terrain\,(no angle information).

\begin{figure}[h]
    \centering
    \includegraphics[width=0.90\linewidth]{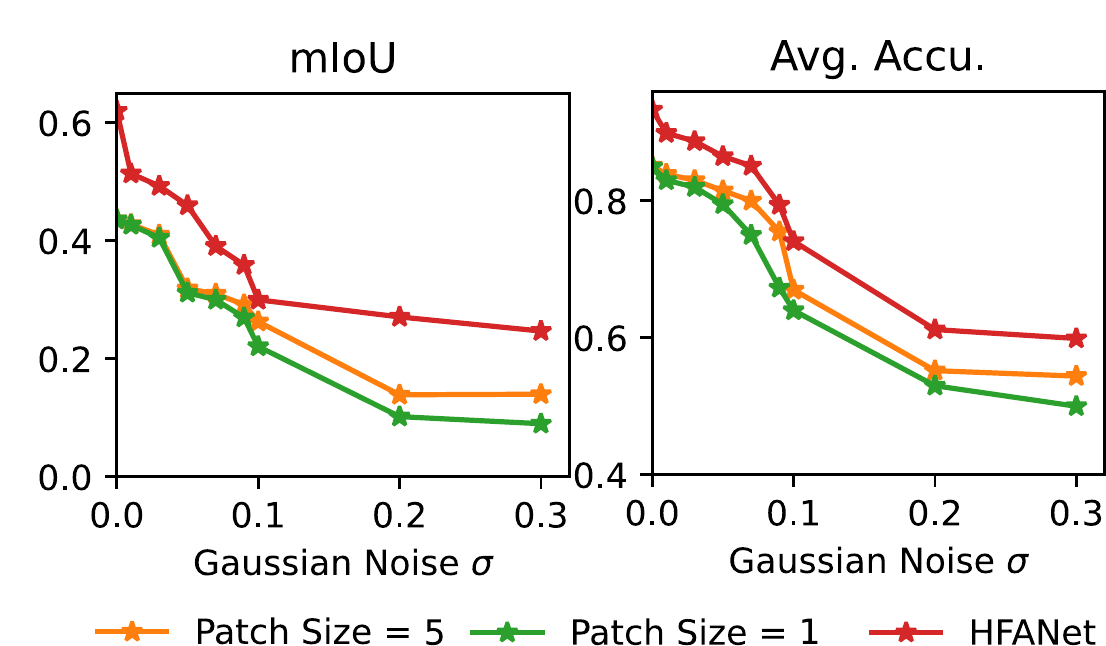}
    \caption{Pose noise test: performance variation trend under various Gaussian pose noise level.}
    \label{fig:pose_noise_compare}
\end{figure}

\begin{table*}[t]
    \small
    \setlength{\tabcolsep}{5pt}
  \caption{Detailed IoU score for each category on Semantic-KITTI\cite{semantic_kitti} dataset}
  \label{semantic-kitti-result1}
  \centering
  
  \begin{tabular}{|c|p{0.7cm}<{\centering}p{0.7cm}<{\centering}p{0.7cm}<{\centering}p{0.7cm}<{\centering}p{0.7cm}<{\centering}p{0.7cm}<{\centering}p{0.7cm}<{\centering}p{0.7cm}<{\centering}p{0.7cm}<{\centering}p{0.7cm}<{\centering}p{0.7cm}<{\centering}p{0.7cm}<{\centering}p{0.7cm}<{\centering}|}
  \hline
  Method &  {road} &  {side-walk} &  {build-ing} &  {fence} &  {pole} &  {traffic-sign} &  {veget-ation} &  {terrain} &  {person} &  {rider} &  {car} &  {motor-cycle} &  {bicycle} \\
  \hline
 PointNet~\cite{pointnet}               & 0.031 & 0.069 & 0.113 & 0.043 & 0.036 & 0.022 & 0.041 & 0.054 & 0.000 & 0.000 & 0.052 & 0.002 & 0.003 \\
 PointNet++~\cite{pointnet_pp}          & 0.066 & 0.023 & 0.079 & 0.042 & 0.112 & 0.014 & 0.036 & 0.183 & 0.000 & 0.002 & 0.133 & 0.010 & 0.000 \\
 RangeNet++(CRF)~\cite{milioto2019iros} & 0.878 & 0.745 & 0.742 & 0.232 & 0.252 & 0.313 & 0.612 & 0.875 & 0.088 & 0.356 & 0.853 & 0.375 & 0.176 \\
 RangeNet++(KNN)~\cite{milioto2019iros} & 0.895 & 0.769 & 0.819 & 0.258 & 0.333 & 0.291 & 0.648 & 0.896 & 0.114 & 0.414 & 0.856 & 0.178 & 0.183 \\
KPConv~\cite{KPConv}                    & 0.738 & 0.574 & 0.653 & 0.244 & 0.469 & \textbf{0.400} & 0.533 & 0.767 & 0.249 & \textbf{0.696} & 0.739 & 0.360 & 0.000 \\
 RandLANet~\cite{RandLA-Net}            & 0.883 & 0.760 & 0.883 & 0.323 & 0.537 & 0.319 & 0.731 & 0.910 & 0.216 & 0.572 & 0.909 & 0.470 & 0.003 \\
  \hline
 Image Based BoF=5              & 0.888 & 0.710 & 0.378 & 0.154 & 0.189 & 0.362 & 0.598 & 0.889 & 0.210 & 0.055 & 0.563 & 0.533 & 0.146 \\
 \hline
 HiFANet                                & \textbf{0.910} & \textbf{0.790} & \textbf{0.903} & \textbf{0.349} & \textbf{0.540} & 0.374 & \textbf{0.755} & \textbf{0.912} & \textbf{0.247} & 0.577 & \textbf{0.933} & \textbf{0.547} & \textbf{0.169} \\
 \hline
  \end{tabular}
\end{table*}

The global comparison of various methods is shown in Fig.\,\ref{fig:quantiative_global}. We can observe that point based method\,(C. RandLA-Net) failed to predict the large traffic sign\,(red box in the RGB image) because such samples are rarely seen in training dataset. At the same time, due to the pose noise, image based method distributes car 3D points to large area\,(see the largely distributed red points in sub-figure D., near the light blue). Our proposed HiFANet can maximally avoid these dilemmas. It obtains semantic representation from RGB images, so it does not require massive training dataset and large presence of all classes. The hierarchical attention design and the involvement of 3D structural prior equip HiFANet with capability to dynamically alleviate the erroneous prediction led by pose noise. In sum, our proposed HiFANet achieves promising performance with relatively small training dataset. It also exhibits pose noise tolerance capability, which is a common challenge in real scenario.

\begin{table}[t]
    \centering
    \caption{HiFANet Network Architecture. FC indicates fully-connection layer, K,Q indicates the key and query in the self-attention module.}
    \begin{tabular}{|c|c|c|}
        \hline
        layer & filter num  & output size \\ 
        \hline
        \multicolumn{3}{|c|}{Input: [B, 10, 5, 5, 5, 256]}\\
        \hline
        \multicolumn{3}{|c|}{Patch Attention Module} \\
        \hline
        K,Q & 64, head num = 4 & [B, 10, 5, 256]\\
        \hline
        FeedForward Net & 256 & [B, 10, 5, 256] \\
        \hline
        \multicolumn{3}{|c|}{Instance Attention Module} \\
        \hline
        K,Q & 64, head num = 4 & [B, 10, 256]\\
        \hline
        FeedForward Net & 256 & [B, 10, 256] \\
        \hline
        \multicolumn{3}{|c|}{InterPoint Attention Module} \\
        \hline
        K,Q & 64, head num = 4 & [B, 10, 256]\\
        \hline
        FeedForward Net & 256 & [B, 10, 256] \\
        \hline
        \multicolumn{3}{|c|}{InterPoint Attention: Structural Prior} \\
        \hline
        FC & 128 & [B, 10, 128]\\
        \hline
        FC & 256 & [B, 10, 256] \\
        \hline
        \multicolumn{3}{|c|}{Classification Head} \\
        \hline
        FC & 512 & [B, 10, 512]\\
        \hline
        FC & 512 & [B, 10, 512] \\
        \hline
        FC & class num & [B, 10, class num]\\
        \hline
    \end{tabular}
    \label{hfa_network}
\end{table}

\subsection{More Experimental Result}

We report the detailed mIoU and mAP score for each individual class in Table\,\ref{semantic-kitti-result1}. We can see from the table that our proposed HiFANet achieves the best performance on most categories. Image based method\,(with BoF=5) obtains inferior performance on some categories such as car, rider and traffic sign, due to the pose noise. Our proposed HiFANet maximally resists the negative impact of pose noise and thus is capable of obtaining promising performance.

\subsection{Discussion on Pose Noise}
We further want to test our proposed HiFANet performance under various pose noise level. To this end, we add Gaussian pose noise to the point-to-image projection matrix in Eqn.\ref{eqn:projection}. The Gaussian noise level is controlled by the Gaussian deviation $\sigma$\,(the mean value is set 0). We compare HiFANet with two image aggregation variants: with patch size 1 and 5. Since the introduction of patch observation is to handle pose noise, it helps us to understand patch observation\,(patch size = 5) resistance to pose noise against the original observation\,(patch size = 1), and against HiFANet.

The Gaussian pose noise $\sigma$ is linearly spaced from 0 to 0.3. The result is shown in Fig.\,\ref{fig:pose_noise_compare}, from which we can observe that adding more pose noise reduces the performance of all methods. The variant with patch size 1 suffers most while HiFANet maximally mitigates the pose noise impact. It thus shows the advantage of involving patch observation in tackling pose noise and our carefully designed HiFANet is capable of learning pose noise tolerant feature representation.

\subsection{Implementation Detail and Source Code}
In HiFANet, we set image observation number as 5, the number of 3D points as 10 and the patch size as 5, so the input size is $[10,5,5,5,256]$. The multi-head attention module head number is 4. The total training dataset is more than 100 million, we randomly subsample 0.5 million points. We implement in PyTorch and train with SGD optimizer, the initial learning rate is 0.1 and decays with factor 0.5 every 30 epochs. Batchsize is 64. The network is trained 100 epochs in total. The network architecture is shown in Table\,\ref{hfa_network}. We use the same hyper-parameter setting to train all other HiFANet variants in our ablation study. For comparing methods, we use their released source code with default or recommended training strategy.

\section{Conclusion and Limitation Discussion}
We propose a three-stage hierarchical fully attentive network, HiFANet, to label the point cloud semantically. The patch observation strategy and bag-of-frames multi-view observation enable HiFANet to handle point-image projection pose noise. Compared to point cloud based methods, HiFANet requires significantly less amount of data and outperforms point based methods by a large margin. The downside our method is that HiFANet's good performance still depends relatively on the LiDAR-camera pose accuracy. If the pose accuracy drops significantly, HiFANet's performance reduces accordingly. Designing more pose-noise tolerant method thus forms a potential future research direction. Another point is that HiFANet only builds on 2D image observations, a joint learning from both the image and point cloud may further improve the performance.

\newpage
\bibliographystyle{plainnat}
\bibliography{references}

\end{document}